# Sub-network Multi-objective Evolutionary Algorithm for Filter Pruning

1st Xuhua Li
*Dept. of Electronics and Information Engineering*
Shenzhen University
Shenzhen, China
lixuhua2021@email.szu.edu.cn

2nd Weize Sun*
*Dept. of Electronics and Information Engineering*
Shenzhen University
Shenzhen, China
proton198601@hotmail.com

3rd Lei Huang
*Dept. of Electronics and Information Engineering*
Shenzhen University
Shenzhen, China
lhuang8sasp@hotmail.com

4th Shaowu Chen
*Dept. of Electronics and Information Engineering*
Shenzhen University
Shenzhen, China
shaowu-chen@foxmail.com

*Abstract*—Filter pruning is a common method to achieve model compression and acceleration in deep neural networks (DNNs). Some research regarded filter pruning as a combinatorial optimization problem and thus used evolutionary algorithms (EA) to prune filters of DNNs. However, it is difficult to find a satisfactory compromise solution in a reasonable time due to the complexity of solution space searching. To solve this problem, we first formulate a multi-objective optimization problem based on a sub-network of the full model and propose a Sub-network Multi-objective Evolutionary Algorithm (SMOEA) for filter pruning. By progressively pruning the convolutional layers in groups, SMOEA can obtain a lightweight pruned result with better performance. Experiments on VGG-14 model for CIFAR-10 verify the effectiveness of the proposed SMOEA. Specifically, the accuracy of the pruned model with 16.56% parameters decreases by 0.28% only, which is better than the widely used popular filter pruning criteria.

## I. INTRODUCTION

In recent years, deep neural networks (DNNs) have been widely applied in various fields due to significant improvement of the computational power brought by graphic processing unit (GPU), which can intelligently solve various complex tasks, such as autonomous driving [1], smart home [2], outlier detection [3] and game option decision [4]. However, DNNs with excellent performance usually have a larger number of parameters, and require expensive computing and memory resources. This problem forbids the DNNs to be deployed on embedded platforms with limited computing and storage capacity, such as mobile phones, when real-time processing of complex information is required. Therefore, it is crucial to compress the model and accelerate the inference procedure.

There are several methods for compressing and accelerating DNNs. The mainstream methods are quantization, low-rank decomposition, knowledge distillation and parameter pruning.

Among them, quantization methods [5]-[7] reduce the quantitative precision of the parameters with the number of parameters unchanged. Different from that, the low-rank decomposition methods [8], [9] decrease the number of parameters by decomposing matrices or tensors into smaller matrices or tensors. Knowledge distillation [10], on the other hand, takes advantage of a teacher network with more parameters to train the student network with fewer parameters.

Generally speaking, the parameter pruning methods can be divided into two categories: weight pruning [11], [12] and filter pruning [13], [14]. The former deletes nodes in neural networks and causes irregular sparsity, which requires specific hardware and software computation libraries to achieve acceleration, while the latter removes filters to produce a structured sparse model and is much easier to be implemented and can accelerate DNNs without customized hardware, making it popular in research and industry in recent years. To prune unimportant filters, pruning algorithms usually rely on heuristic criteria, including scale-based and relationship-based ones. Scale-based criteria judge the importance of filters by their magnitudes, such as $l_1$ norm [15] and $l_2$ norm [16], and prune filters with minor magnitudes. Relationship-based criteria consider the correlativity of filters [17] and prune the most replaceable filters.

In recent years, some scholars have used multi-objective optimization algorithms [18] for model compression. The number of parameters and accuracy of DNNs are considered as two constrained optimization objectives, then the evolutionary algorithms (EA) can be applied to find a compromise solution. In [19], EA was used for finding a sparse and low-rank model, and [20], [21] utilized EA to balance the classification accuracy and compression ratio for filter pruning. However, these methods took classification accuracy as one of the objectives in EA and pruned all layers of one network simultaneously. Since the search space of a deep neural network with a large number of convolutional layers is huge, such approaches become very time-consuming to find a satisfactory solution, forbidding them to be applied in large networks.

In order to accelerate the searching time of EA and prune filters of DNNs without severe performance degradation, we

The work described in this paper was supported in part by the Foundation of Shenzhen under Grant JCYJ20190808122005605, and in part by the Guangdong Basic and Applied Basic Research Foundation under Grant 2021A1515011706, and in part by the National Natural Science Foundation of China (NSFC) for Grant 62101335, 61925108 and U1913203.
* Corresponding author.

propose a Sub-network Multi-objective Evolutionary Algorithm (SMOEA) for filter pruning. Different from existing EA-based filter pruning methods that directly optimize the final classification accuracy, SMOEA prunes a layer by considering minimizing reconstruction errors of output feature maps in the next layer. In this manner, SMOEA takes two-layer blocks as the minimum units for evolution process, and can optimize a full network by optimizing one or several blocks gradually. By group-based progressive pruning, SMOEA can find the solution for each step greedily and finally, attain a satisfactory full-layers solution with better performance.

## II. METHODOLOGY

### A. Notations and Definitions

A deep neural network is defined as $f(X;W)$, where $X$ is input data. The $W = \{W_l\}_{l=1}^L$ denotes the network parameters where $L$ is the number of layers and $W_l$ is the network parameters of the $l$-th layer. The $W' = \{W'_l\}_{l=1}^L$ is defined as the pruning result of the network $W$. Generally speaking, the $W'_l$ can be obtained from $W_l$ by

$$W'_l = W_l \odot M_l, \quad (1)$$

where $M_l$ denotes the binary mask of the $l$-th layer in which 0 and 1 indicate the filter being pruned and retained, respectively.

**The $l$-th sub-network.** $subf_l(Map_l;\cdot)$ represents the $l$-th sub-network. Specifically, that is a two-layer block consisting of the $l$-th convolutional layer $W_l$ to be pruned and the $(l+1)$-th layer $W_{l+1}$. $Map_l$ denotes the input feature maps of the $l$-th layer or the $l$-th sub-network, and $\cdot$ is $\{W_l, W_{l+1}\}$ before pruning or $\{W_l \odot M_l, W_{l+1}\}$ after pruning. As illustrated in Fig. 1, unimportant filters of the first convolutional layer in the $l$-th sub-network, i.e., slices of $W_l$, will be pruned, while the second layer $W_{l+1}$ is used for feature maps extraction here only but will be pruned in the $(l+1)$-th sub-network.

**Population and elite offspring.** In the evolution process, we denote the $t$-th generated population with $N$ individuals as $\{P_n^t\}_{n=1}^N$. Similarly, $\{P'^t_k\}_{k=1}^K$ refers to the $t$-th generated elite offspring with $K$ individuals.

### B. Definition of Optimization Problem

In filter pruning, the target is to prune as many filters as possible to reduce the number of network parameters under satisfying performance. Therefore, it can be regarded as a multi-objective optimization problem: good performance and fewer parameters. However, taking classification accuracy as the performance objective will lead to an extremely complex solution space. Therefore, we define $Error_l$, the reconstruction error of output feature maps in the $l$-th sub-network, as the objective. Then the optimization problem can be defined as follows:

$$\min_{M_l} F(M_l) = \left(obj_1(M_l), obj_2(Map_l, \{W_l, W_{l+1}\}, M_l)\right)^T$$
$$= (Filter\%_l, Error_l)^T$$
$$s.t. \quad \tau_1 \leq Filter\%_l \leq \tau_2, \quad (2)$$

where $Filter\%_l$, i.e., $obj_1$, represents the proportion of remained filters in the $l$-th convolutional layer $Conv_l$ after pruning:

$$Filter\%_l = \frac{\|M_l\|_0}{\#(M_l)}, \quad (3)$$

where # mean "the total number of".

The calculation of $Error_l$, i.e., $obj_2$, is defined as:

$$Error_l = \|Map_{l+2} - \alpha\widetilde{Map}_{l+2}\|_2, \quad (4)$$

where

$$Map_{l+2} = subf_l(Map_l; \{W_l, W_{l+1}\}) \quad (5)$$
$$\widetilde{Map}_{l+2} = subf_l(Map_l; \{W_l \odot M_l, W_{l+1}\}) \quad (6)$$
$$\alpha = \text{argmin}_\alpha \|Map_{l+2} - \alpha\widetilde{Map}_{l+2}\|_2 \quad (7)$$

$Map_{l+2}$ and $\widetilde{Map}_{l+2}$ represent the output feature maps in the $l$-th sub-network before and after pruning its first layer, respectively, as shown in Fig. 1. Intuitively, the energy of $\widetilde{Map}_{l+2}$ after pruning is smaller than $Map_{l+2}$ without pruning, therefore, we compensate the former by introducing an intensity value $\alpha$ and get $\alpha \times \widetilde{Map}_{l+2}$.

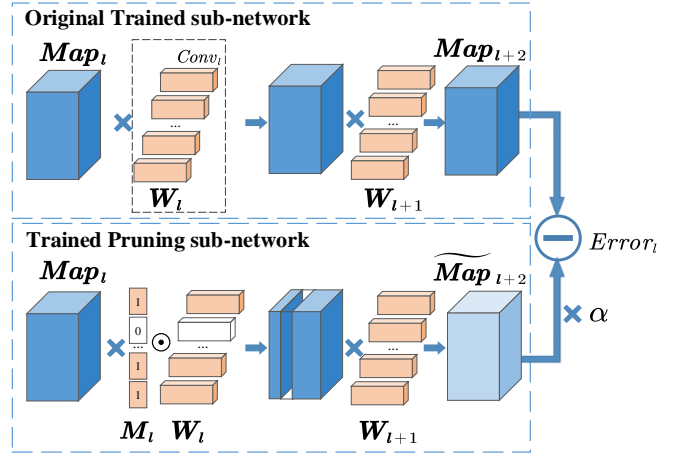

Fig. 1. Pruning filters of $Conv_l$ in the $l$-th sub-network.

### C. Solution

We use NSGA-II [22] to solve the optimization problem (2) of the $l$-th sub-network for $l = 1,2,...L$ independently as in Algorithm 1. The parameters $\{W_l, W_{l+1}\}$ and the feature maps $Map_l$ are the input of the algorithm. The $M_l$ is an individual solution from the final elite offspring and $Filter\%_l$ is limited between $\tau_1$ and $\tau_2$. Note that $Filter\%_l$ and $Error_l$ as two attributes of the individual are related to sorting and selecting elite offspring.

The whole evolutionary process of Algorithm 1 is divided into five major parts:

*a)* **Initialization**: Initialize the original generation $\{P_n^0\}_{n=1}^N$ with $N$ individuals. The individuals are generated uniformly in $[\tau_1, \tau_2]$ with random selection of pruned filters.

*b)* **Sorting**: The individuals of the population are sorted by non-dominated solution and crowding distance successively

[22], and $K$ outstanding individuals are selected as elite offspring $\{P_k^0\}_{k=1}^K$.

*c)* **Generating**: The $(t+1)$-$th$ generation $\{P_n^{t+1}\}_{n=1}^N$ are generated from the $t$-$th$ elite offspring $\{P_k'^t\}_{k=1}^K$. Each individual of $\{P_n^{t+1}\}_{n=1}^N$ is generated by performing genetic operation including crossover and variation [22] on two randomly selected individuals of $\{P_k'^t\}_{k=1}^K$.

*d)* **Sorting Elite offspring**: $\{P_n^{t+1}\}_{n=1}^N$ and $\{P_k'^t\}_{k=1}^K$ are combined to form the $(t+1)$-$th$ generation population. $K$ individuals of them are selected to be the elite offspring $\{P_k'^{t+1}\}_{k=1}^K$ of next generation according to b).

We perform c) and d) iteratively as the process of offspring evolution until a maximum number of generations $T$ is reached.

*e)* **Finding**: The final step is to find a representative individual. The Pareto Front for the last generation of the elite offspring $\{P_k'^T\}_{k=1}^K$ can be drawn and the knee point farthest from the normal line segment of the edge endpoints is selected as the representative individual $M_l$.

---
**Algorithm 1** Evolution Process
---
**Input:** $Map_l$, $\{W_l, W_{l+1}\}$, $t = 0$
**Output:** $M_l$
1: $Initialize\ \{P_n^0\}_{n=1}^N$
2: $\{P_k'^0\}_{k=1}^K \leftarrow Sort\ \{P_n^0\}_{n=1}^N$
3: **while** $t < T$ **do**
4:   $\{P_n^{t+1}\}_{n=1}^N \leftarrow Generate\ from\ \{P_k'^t\}_{k=1}^K$
5:   $\{P_k'^{t+1}\}_{k=1}^K \leftarrow Sort\ \{\{P_n^{t+1}\}_{n=1}^N \cup \{P_k'^t\}_{k=1}^K\}$
6:   $t \leftarrow t + 1$
7: **end while**
8: $M_l \leftarrow Find\ from\ \{P_k'^T\}_{k=1}^K$
---

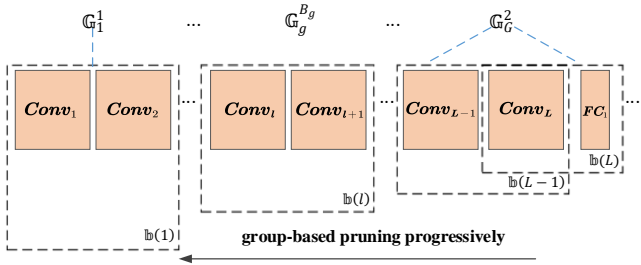

Fig. 2. Pruning filters in groups successively, in which one group contains one or more blocks.

*D. Main Framework of SMOEA*

As shown in Fig.2, assume that there are totally $L$ continuous convolutional layers to be pruned and define the $l$-th sub-network as the $l$-th block $\mathbb{b}(l)$. We further combine one or more consecutive blocks to form a group. The $g$-th group including $B_g$ blocks, is represented as:

$$\mathbb{G}_g^{B_g} = \{\mathbb{b}(l)|l = l_0 - 1 + \sum_{i=1}^{g-1} B_i + b_g\ ;\ b_g = 1,2,...,B_g\}, (8)$$

where $l_0$ is the index of the first block to be pruned. There are G groups totally: $\{\mathbb{G}_g^{B_g}|\ g = 1,2,...,G\}$. Note that $\sum_{i=1}^G B_i = L$. An example of the relationship between the block and the group is shown in Fig.2. Here we set $l_0 = 1$ in (8), then the $\mathbb{G}_G^2$ represents the $G$-th group and contains 2 blocks $\mathbb{b}(L-1)$ and $\mathbb{b}(L)$.

The detailed steps of the proposed Sub-network Multi-objective Evolutionary Algorithm (SMOEA) for filter pruning are described in Algorithm 2. Given training set $X$, $\{W_l\}_{l=1}^L$ of a well-trained network and a group set $\{\mathbb{G}_g^{B_g}|g = 1,2,...,G\}$, we can obtain the pruned network by performing algorithm 1 on each $\mathbb{G}_g^{B_g}$ and fine-tuning progressively. Note that different from one shot pruning, the group $\mathbb{G}_g^{B_g}$ is the minimum unit for pruning. To avoid error accumulation, we prune groups progressively in a reversed order, i.e., the $G$-th group is pruned first while the first group is pruned in the end. When pruning a group $\mathbb{G}_g^{B_g}$, algorithm 1 is used to find $M_l$ for pruning $W_l$ independently. After pruning the group, fine-tuning is followed to recover the performance. The group-wise pruning and fine-tuning can be carried out iteratively for $G$ times then the final compressed network is obtained.

---
**Algorithm 2** Main Framework of SMOEA
---
**Input:** $X$, $\{W_l\}_{l=1}^L$ & $\{\mathbb{G}_g^{B_g}|g = 1,2,...,G\}$
**Output:** $\{W'_l\}_{l=1}^L$
1: **for** $(g = G, G-1, ..., 1)$ **do**
2:   **for** $b_g = 1,2,...,B_g$ **do**
3:     $Calculate\ M_l\ from\ (Map_l, \{W_l, W_{l+1}\})$ according to Algorithm 1
4:     $W'_l \leftarrow W_l \odot M_l$
5:   **end for**
6:   $\{W'_l\}_{l=1}^L \leftarrow Finetune(f(X, \{W'_l\}_{l=1}^L))$
7:   $\{W_l\}_{l=1}^L \leftarrow \{W'_l\}_{l=1}^L$
8: **end for**
---

### III. EXPERIMENTAL RESULTS AND ANALYSIS

In this section, we first evaluate the rationality of the proposed $a$ to find knee point of Pareto Front and then verify the effectiveness of the proposed SMOEA in selecting filters.

*A. Experimental Settings*

The experiment dataset is CIFAR-10, which consists of 60,000 color images with dimensions $32 \times 32$ from 10 categories. There are 50,000 images in the training set and 10,000 images in the test set. The VGG-14 model, which is composed of 13 convolutional layers for feature extraction and a classification FC layer with 512 input nodes and 10 output nodes, is to be pruned. VGG-14 is modified from the well-known VGG-16 [23] by replacing the final 3 FC layers with one classification FC layer.

In the evolution process, the number of individuals of each generation population is $N = 100$, the number of individuals of each elite offspring is $K = 30$ and the maximum number of generations is T = 100. The probabilities of crossover and mutation are 1 and 0.05, respectively, ensuring the population's heritability and evolvability. The scope of the $Filter\%_i$ constraints for each $Conv_i$ is $[\tau_1, \tau_2] = [0.2, 0.8]$.

In the fine-tuning process, the initial learning rate is set as $lr = 0.01$, and the total number of training epochs is 160. In order to achieve better convergence of the model, $lr$ is divided by 10 in the 50-th epoch and the 100-th epoch, respectively. To avoid the randomness of the experiment, the experiment is repeated three times to get the average accuracy for comparison.

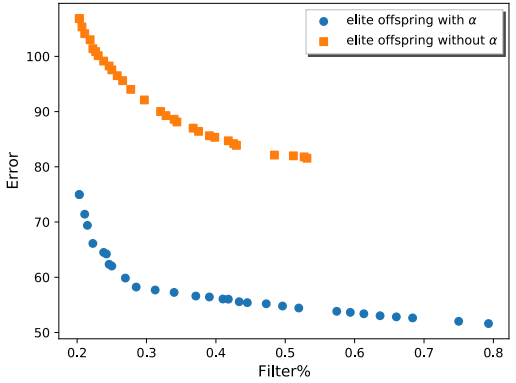

(a) Pareto Front of $\mathbb{b}(5)$

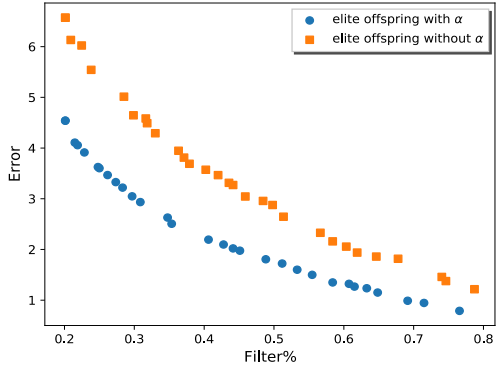

(b) Pareto Front of $\mathbb{b}(10)$

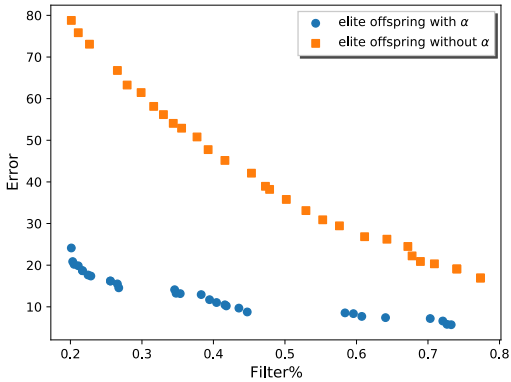

(c) Pareto Front of $\mathbb{b}(13)$

Fig. 3. Pareto Fronts of the last generation of elite offspring with $\alpha$ and without $\alpha$. Subfigure (a–c) refer to the Pareto Fronts of $\mathbb{b}(5)$, $\mathbb{b}(10)$ and $\mathbb{b}(13)$, respectively.

### B. The Rationality of $\alpha$ in Evolution Process

The distributions of final Pareto Front from Algorithm 1 on $\mathbb{b}(5)$, $\mathbb{b}(10)$ and $\mathbb{b}(13)$ are first shown. Two cases, referred to as 'with $\alpha$' and 'without $\alpha$' will be tested and compared to shown the rationality of the parameter $\alpha$ in Algorithm 1. Specifically, 'with $\alpha$' refers to generate the elite offspring together with optimizing $\alpha$ in equation (7), while 'without $\alpha$' means setting $\alpha = 1$ directly. The results are shown in Fig. 3, in which each point refers to an individual of the last generation of the elite offspring, with $Filter\%$ and $Error$ being the x and y axis, respectively. It is shown that by optimizing $\alpha$, the Pareto Front obtained is more inclined to the lower left with a greater degree of bending compared 'without $\alpha$' one. Furthermore, according to Fig. 3(a), it can be inferred that the 'with $\alpha$' approach can avoid losing the elite individuals in some specific range of $Filter\%$, showing the ability of generating a better final Pareto Front.

### C. Experiments on VGG14

To demonstrate the effectiveness of SMOEA on pruning large-scale DNNs, we prune the VGG-14 network for the CIFAR-10 dataset. We first train VGG-14 as a benchmark on the training set to achieve an accuracy of 93.79%. Groups are set as $\{\mathbb{G}_1^1, \mathbb{G}_2^1, \mathbb{G}_3^1, \mathbb{G}_4^6\}$ with $l_0 = 5$, which is, only $Conv_5$ to $Conv_{13}$ are pruned as the number of parameters in other layers is small.

The results are shown in TABLE I. Compared with the original full-size model, the model pruned by SMOEA contains only 16.56% parameters and 42.17% FLOPs, however, its accuracy is only slightly worse than that of the original model.

We also compare the SMOEA with random pruning (rand) and two widely used filter pruning criteria: $l_2$ [15] and FPGM [16]. Note that the former follows scale-based criterion while the latter follows relationship-based criterion. Two kinds of retained rates $\{Filter\%_l\}_{l=1}^L$ are applied to the rand, $l_2$ and FPGM methods as written in TABLE I: the ones without '-fix' refer to the methods pruning the layers with the same retained rates as those of the SMOEA while the ones with '-fix' refer to the methods pruning the $Conv_5$ to $Conv_{13}$ layers under a fixed retained rates making the final percentage of total remained parameter equals to the former approximately.

It is shown in TABLE I that under the same retained rates of all the layers, rand, $l_2$ and FPGM are inferior than SMOEA (93.34%, 93.19% and 93.31% vs 93.51%), demonstrating the superior performance in finding the unimportant filters of SMOEA. It also indicates the strong performance in finding appropriate pruning rate between different layers. According to TABLE I, the rand, $l_2$ and FPGM are all superior to rand-fix, $l_2$-fix and FPGM-fix, respectively with approximated the same global percentage of remained parameter 16.56%. This shows that the SMOEA is good at searching appropriate network structures.

TABLE I. PRUNING RESULTS OF VGG-14 ON CIFAR-10

| Model | Method | Remained parameter% | FLOPs | Accuracy% |
|---|---|---|---|---|
| VGG-14 | original | - | 6.26E+08 | 93.79 |
| | rand-fix[a] | 16.58 | 2.65E+08 | 93.24($\pm$0.08) |
| | $l_2$-fix | | | 93.12($\pm$0.13) |

| Model | Method | Remained parameter% | FLOPs | Accuracy% |
|---|---|---|---|---|
| | FPGM-fix | | | 93.05($\pm$0.04) |
| | rand | | | 93.34($\pm$0.03) |
| | $l_2$ | 16.56 | 2.64E+08 | 93.19($\pm$0.13) |
| | FPGM | | | 93.31($\pm$0.03) |
| | SMOFP(Ours) | | | **93.51($\pm$0.05)** |

a. "-fix": each convolutional layer is pruned under the same retained rate

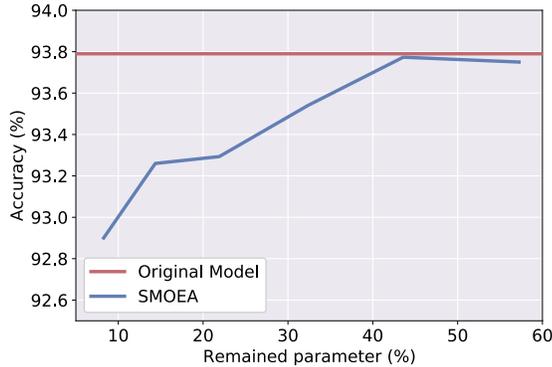

Fig. 4. Accuracies of VGG-14 for CIFAR-10 with different proportion of remained parameters, including six pruned models with pruning strategy of similar $Filter\%$ from 0.25 to 0.75 and an interval of 0.1.

In the end, the performance of the SMOEA approach under the different percentage of remained parameters is tested. In this case, all the layers will be pruned under the same or similar $Filter\%$ selected from the final elite offspring from Algorithm 1. The accuracies of pruned models under different percentages of remained parameters are shown in Fig. 4. When 43% of parameters are retained, the pruned model can achieve the same accuracy as the full model before pruning, showing the effectiveness of the proposed evolutionary method. Furthermore, to achieve an accuracy of 93.51%, about 30% of parameters should be retained when remaining parameters are set to be similar for all pruned layers. Using the knee point of each pruned layer, on the other hand, required only 16.56% of remaining parameters to get the same accuracy, further showing the efficiency of the proposed method.

IV. CONCLUSION

In this paper, a Sub-model Multi-objective Evolutionary Algorithm (SMOEA) for filter pruning is proposed. Unlike traditional evolutionary algorithms that take classification accuracy as one of the optimization objectives, SMOEA focus on minimizing the reconstruction error of output feature maps in sub-networks progressively, and can effectively reduce the complexity of the evolution process. Therefore, SMOEA can efficiently delete redundant filters to compress and accelerate DNNs, and also find the optimal structures by exploring pruning rates between different layers. Experimental results show that SMOEA can prune more than 83% of parameters in VGG-14 without significant loss in accuracy.

REFERENCES

[1] Y. P. Pan, C. A. Cheng and K. Saigol, "Agile autonomous driving using end-to-end deep imitation learning," arXiv:1709.07174, 2017.
[2] V. Bianchi, M. Bassoli, G. Lombardo, P. Fornacciari, M. Mordonini and I. De Munari, "IoT Wearable Sensor and Deep Learning: An Integrated Approach for Personalized Human Activity Recognition in a Smart Home Environment," in IEEE Internet of Things Journal, vol. 6, pp. 8553-8562, 2019.
[3] W. N. Lu, Y. Cheng, C. Xiao, S. Y. Chang, S. Huang and B. Liang, "Unsupervised sequential outlier detection with deep architectures," IEEE transactions on image processing, vol. 26, pp. 4321-4330, 2017.
[4] S. David, S. Julian, S. Karen et al, "Mastering the game of go without human knowledge," Nature, vol. 550, pp. 354-359, 2017.
[5] M. Courbariaux, Y. Bengio and J. David, "BinaryConnect: Training Deep Neural Networks with binary weights during propagations," Advances in neural information processing systems 28, pp. 3123-3131, Quebec, Canada. 2015.
[6] I. Hubara, M. Courbariaux, D. Soudry, R. El-Yaniv and Y. Bengio, "Binarized neural networks," Advances in neural information processing systems 29, pp. 4107–4115, Barcelona, Spain, 2016.
[7] M. Rastegari, V. Ordonez, J. Redmon and A. Farhadi, "Xnornet: Imagenet classification using binary convolutional neural networks," arXiv:1603.05279, 2016.
[8] S. W. Chen, J. H. Zhou, W. Z. Sun and L. Huang, "Joint Matrix Decomposition for Deep Convolutional Neural Networks Compression," arXiv:2107.04386, 2021.
[9] W. Z. Sun, S. W. Chen, H. C. So and M. Xie, "Deep Convolutional Neural Network Compression via Coupled Tensor Decomposition," IEEE Journal of Selected Topics in Signal Processing, vol.15, pp.603-616, 2021.
[10] G. Hinton, O. Vinyals and J. Dean, "Distilling the Knowledge in a Neural Network", arXiv:1503.02531, 2015.
[11] Y. LeCun, J. Denker and S. Solla, "Optimal brain damage," Advances in Neural Information Processing Systems 2, pp. 598-605, Colorado, USA, 1989.
[12] S. Han, J. Pool, J. Tran and W. Dally, "Learning both Weights and Connections for Efficient Neural Network," Advances in Neural Information Processing Systems 28, pp. 1135-1143, Quebec, Canada, 2015.
[13] Y. He, X. Y. Dong, G. L. Kang, Y. W. Fu, C. G. Yan, and Y. Yang, "Asymptotic soft filter pruning for deep convolutional neural networks," IEEE transactions on cybernetics, vol. 50, pp. 3594-3604, 2019.
[14] X. D. Wang, Z. D. Zheng, Y. He, F. Yan, Z. Q. Zeng and Y. Yang, "Progressive local filter pruning for image retrieval acceleration," arXiv:2001.08878, 2020.
[15] H. Li, A. Kadav, I. Durdanovic, H. Samet and H. P. Graf, "Pruning filters for efficient convnets," arXiv:1608.08710, 2016.
[16] Y. He, G. L. Kang, X. Y. Dong, Y. W. Fu and Y. Yang, "Soft filter pruning for accelerating deep convolutional neural networks," arXiv:1808.06866, 2018.
[17] Y. He, P. Liu, Z. W. Wang, Z. L. Hu and Y. Yang, "Filter Pruning via Geometric Median for Deep Convolutional Neural Networks Acceleration," Proceedings of the IEEE/CVF Conference on Computer Vision and Pattern Recognition (CVPR), pp. 4340-4349, Long Beach, California, 2019.
[18] Q. F. Zhang and H. Li, "MOEA/D: A multiobjective evolutionary algorithm based on decomposition," IEEE Transactions on evolutionary computation, vol. 11, pp. 712-731, 2007.
[19] J. H. Huang, W. Z. Sun and L. Huang, "Deep neural networks compression learning based on multiobjective evolutionary algorithms," Neurocomputing, vol. 378, pp. 260-269, 2020.
[20] T. Wu, J. Shi, D. Zhou, X. Zheng and N. Li, "Evolutionary multiobjective one-shot flter pruning for designing lightweight convolutional neural network," Sensors, vol. 21, pp. 5901-5921, 2021.
[21] Y. Zhou, G. G. Yen and Z. Yi, "A Knee-Guided Evolutionary Algorithm for Compressing Deep Neural Networks," In IEEE Transactions on Cybernetics, vol. 51, pp. 1626-1638, 2021.
[22] K. Deb, A. Pratap, S. Agarwal and T. Meyarivan, "A fast and elitist multiobjective genetic algorithm: NSGA-II," IEEE transactions on evolutionary computation, vol. 6, pp.182-197, 2002.
[23] K. Simonyan and A. Zisserman, "Very deep convolutional networks for large-scale image recognition," arXiv:1409.1556, 2014.